\documentclass[10pt,twocolumn,letterpaper]{article}

\usepackage{iccv}
\usepackage{times}
\usepackage{epsfig}
\usepackage{graphicx}
\usepackage{amsmath}
\usepackage{amssymb}

\usepackage{booktabs}
\usepackage{url}
\usepackage{graphicx}
\usepackage[skip=3pt]{subcaption}
\usepackage{stfloats}
\usepackage{amsfonts}
\usepackage[ruled,vlined,linesnumbered]{algorithm2e}
\usepackage[noend]{algpseudocode}
\usepackage{fancyhdr}

\usepackage{xcolor}

\algnewcommand{\IfThenElse}[3]{
  \State \algorithmicif\ #1\ \algorithmicthen\ #2\ \algorithmicelse\ #3}

\usepackage{floatrow}
\newfloatcommand{capbtabbox}{table}[][\FBwidth]

\usepackage{float}
\usepackage{placeins}

\newcommand{\sucr}[4]{ #1$\pm$#2 (#3$\pm$#4) }

\usepackage[pagebackref=true,breaklinks=true,letterpaper=true,colorlinks,bookmarks=false]{hyperref}

\iccvfinalcopy 

\def\iccvPaperID{****} 

\ificcvfinal\fi

\begin{document}

\title{Towards Learning Multi-agent Negotiations via Self-Play}

\author{Yichuan Charlie Tang\\
Apple Inc.\\
{\tt\small yichuan\_tang@apple.com}
}

\maketitle
\ificcvfinal\thispagestyle{empty}\fi

\begin{abstract}
Making sophisticated, robust, and safe sequential decisions is at the heart of intelligent systems. This is especially critical for planning in complex multi-agent environments, where agents need to anticipate other agents' intentions and possible future actions. Traditional methods formulate the problem as a Markov Decision Process, but the solutions often rely on various assumptions and become brittle when presented with corner cases. In contrast, deep reinforcement learning (Deep RL) has been very effective at finding policies by simultaneously exploring, interacting, and learning from environments. Leveraging the powerful Deep RL paradigm, we demonstrate that an iterative procedure of self-play can create progressively more diverse environments, leading to the learning of sophisticated and robust multi-agent policies. We demonstrate this in a challenging multi-agent simulation of merging traffic, where agents must interact and negotiate with others in order to successfully merge on or off the road. 
While the environment starts off simple, we increase its complexity by iteratively adding an increasingly diverse set of agents to the agent ``zoo" as training progresses. Qualitatively, we find that through self-play, our policies automatically learn interesting behaviors such as defensive driving, overtaking, yielding, and the use of signal lights to communicate intentions to other agents. In addition, quantitatively, we show a dramatic improvement of the success rate of merging maneuvers from 63\% to over 98\%.
\end{abstract}

\section{Introduction}\vspace{-0.05in}

\begin{figure}[t]
\begin{center}
\includegraphics[width=1\linewidth]{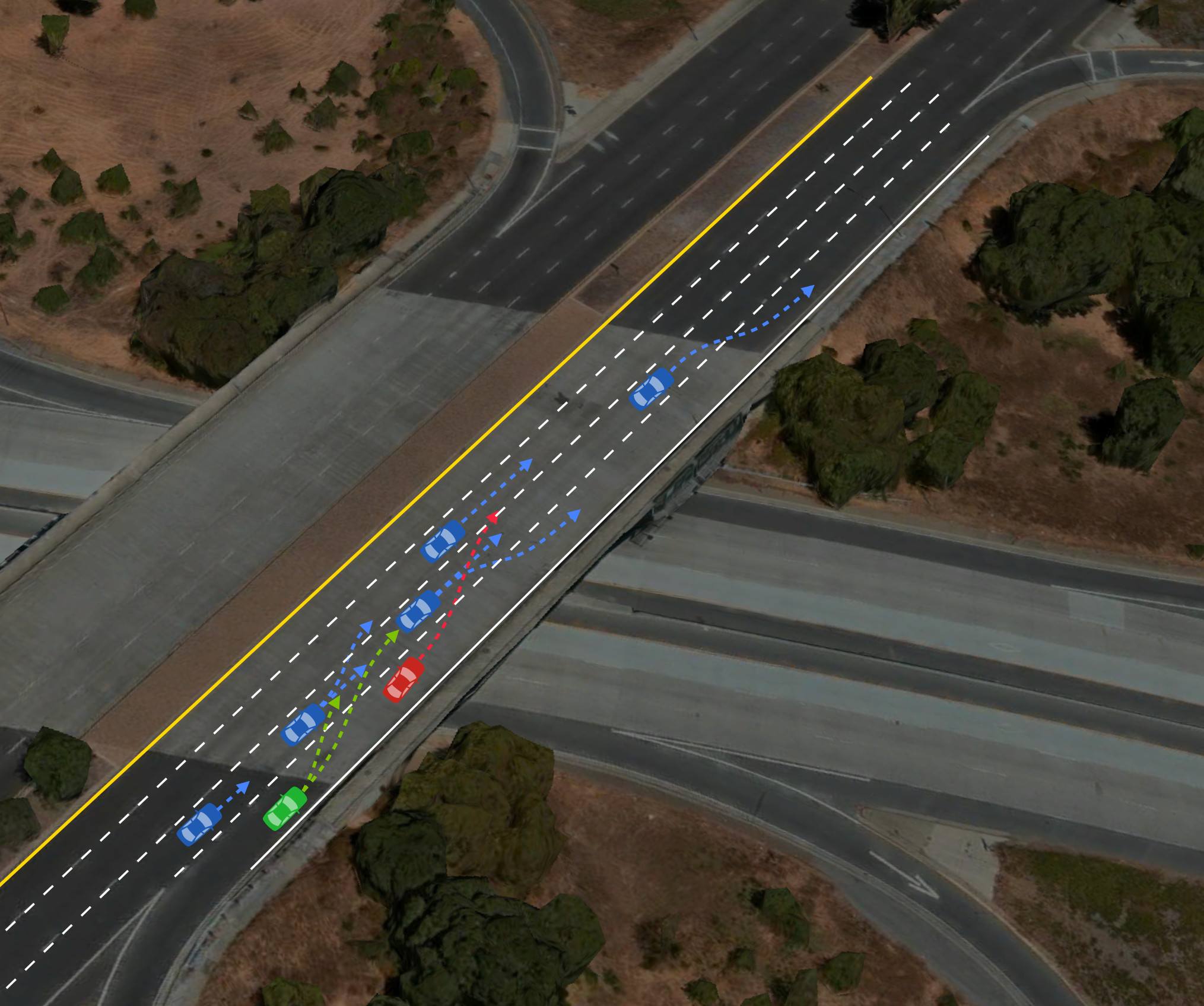}
\end{center}
\vspace{-0.1in}
\caption{ \textit{A merge scenario: the green and red vehicles want to find a gap and merge on, while some of the blue vehicles want to merge off. For robust decision making, it is critical to negotiate with other vehicles on the road}.}
\label{fig:overview}
\vspace{-0.1in}
\end{figure}

One of the key challenges for building intelligent systems is learning to make safe and robust sequential decisions in complex environments. In a multi-agent setting, we must learn sophisticated policies with negotiation skills in order to accomplish our goals. Fig.~\ref{fig:overview} provides an example of this in the domain of autonomous driving, where we wish to learn a policy to safely merge onto the highway in the presence of other agents. Merges are considered complex~\cite{wei2013autonomous,dong2017intention,shalev2016safe}, where behavior planning must accurately predict each other's intentions and try to act rationally in a general sum game~\cite{bucsoniu2010multi}.
Traditional solutions often tackle these Markov Decision Processes (MDPs) by making many assumptions and engineering hard-coded behaviors, often leading to constrained and brittle policies~\cite{hwang2005modeling, Paden16, gonzalez2016review}. For example, in lane changes and merges, a typical approach would first check for a sufficient gap in the adjacent lane (high-level behavior planning), followed by solving for the optimal future trajectory (low-level motion planning). While this is a sensible approach for the majority of situations, it quickly becomes hard to handle various corner cases: e.g. multiple vehicles are simultaneously trying to merge to the same lane. Moreover, it is difficult to reason about how others will react to our own actions and how we might react to their reactions and so forth.

In a drastically different approach to solving the problem, reinforcement learning (RL) can directly learn policies through repeated interactions with its environment. Ideally, the simulated environment should be diverse and realistic to facilitate real world transfers. The policy function is usually a general purpose deep neural network, making no \emph{a priori} assumptions about the possible solutions. Recently, with the increase in computation power and methodological improvements, deep reinforcement learning (deep RL) has been applied to a variety of MDP problems and has achieved striking breakthroughs: superhuman performances in video games~\cite{mnih2016asynchronous}, Go~\cite{silver2017mastering}, and most recently in StarCraft~\cite{alphastarblog}. However, it is unclear whether or not these techniques can be easily transferred to driving applications. Unlike in games, driving is not a zero-sum game and involves partially observed stochastic environments, where safety and following traffic laws are paramount.

In this paper, we leverage deep RL and propose an iterative \emph{self-play} training scheme to learn robust and capable policies for a multi-agent merge scenario in a simulated traffic environment. We first build a simulation of traffic based on real road geometry, where the road network is annotated by aligning with a real ``zipper" merge from satellite imagery, see Fig.~\ref{fig:overview}. We populate the world densely with rule-based agents that are capable of lane-following and safe lane changes. 
However, it quickly becomes apparent that simple rule-based policies are insufficient to deal with all of the complexities of the environment. We can do better by using RL to train policies in the presence of the basic rule-based agents. However, the RL policy easily overfits to the distribution of behaviors of the basic rule-based agents.

To overcome this, we devise an iterative self-play algorithm where previously trained RL policies are mixed with rule-based agents to create a diverse population of agents. In self-play, the policies being learned are also simultaneously used to control agents in the simulation. As training progresses, our agents' capabilities also evolve, increasing the diversity and complexity of the environment. In such an complex environment, a capable policy must learn when to slow down, when to accelerate and pick the gap to merge into. It must also learn to communicate intentions to other agents via turn signals or through its observable behaviors. Lastly, it needs to estimate the latent goals and beliefs of the other agents. Towards learning such policies, we demonstrate the qualitative and quantitative behaviors or our self-play trained policies in Sec.~\ref{sec:experiments}.

While we start with only a particular merge scenario, our underlying motivation is that once we learn good policies for all of the challenging scenarios, driving from any location A to any other location B can be `stitched' together from these scenario specific policies, much like how options are used in hierarchical reinforcement learning. Our main contributions are as follows:

\vspace{-0.1in}
\begin{itemize}
\setlength\itemsep{0.03em}
\item A fast 2D simulation environment with realistic topology and a kinematics Bicycle motion model.
\item Rule-based agents with basic driving abilities: avoiding collisions, making safe lane changes, using turn signal lights.
\item Model-free RL policy learning from a combination of rasterized state encodings as well as low-dimensional state variables.
\item Quantitatively, we demonstrate that we can improve the rate of successful merges from 63\% to 98\% via our proposed self-play strategy.
\item Qualitatively, our self-play trained agents learned human like behaviors such as defensive driving, yielding, overtaking, and using turn signals.
\end{itemize}

\section{Related Works}\label{sec:related_works}\vspace{-0.05in}
Using reinforcement learning algorithms to solve multi-agent systems is useful in a wide variety of domains, including robotics, computational economics, operations research, and autonomous driving. A comprehensive overview and survey on existing multi-agent reinforcement learning (MARL) algorithms is provided by~\cite{bucsoniu2010multi}. More recently, an increase of interest in MARL has come from the perspective of achieving general intelligence, where agents must learn to interact and communicate with each other in a shared environment~\cite{lanctot2017unified,foerster2016learning}. Game theoretic 
approaches for MARL include fictitious play~\cite{heinrich2015fictitious} and fictitious self-play, the latter of which has been recently proposed with deep neural networks for imperfect information poker games and shown to converge to approximate Nash equilibria~\cite{heinrich2016deep}.

Policy learning for the driving domain can be formulated as a MARL problem where each agent is a vehicle and the environment is the road scene. The most challenging aspect of this application domain lies in the ability to be robust and safe~\cite{shalev2017formal}. To this end, a method was proposed to improve functional safety by decomposing the policy function into two components, a learnable function which is gated by a hard constraint, a non-learnable trajectory planner~\cite{shalev2016safe}.

There have also been previous work directly applying deep reinforcement learning to driving simulations. Deterministic actor critic \cite{LillicrapHPHETS15} has been previously applied to control the steering and acceleration in a racing game~\cite{TORCS}, where the policy mapped pixels to actions. As part of the CARLA simulator, the authors also released a set of deep RL baselines results~\cite{Dosovitskiy17}. Deep RL was utilized for an end-to-end lane keep assist system~\cite{sallab2016end}, and also used for navigating a roundabout intersection~\cite{chen2019model}. For related but non-RL approaches, conditional imitation learning agents have also been learned via behavior cloning from expert data~\cite{codevilla2018end}.

\section{Preliminary}\label{sec:background}\vspace{-0.05in}
Before diving into the details of our algorithms and framework, we first provide a background on deep reinforcement learning and its approach to learning optimal policies. We will use the \emph{Markov Decision Process} (MDP) framework to model our problem~\cite{Puterman94}. Reinforcement learning consists of a type of algorithm for solving MDPs in which the agent repeatedly interacts with a stochastic environment by executing different actions. The MDP consists of a state space $\mathcal{S}$, an action space $A$, and a reward function $r(s,a): \mathcal{S} \times \mathcal{A} \rightarrow \mathbb{R}$. The model of the environment is: $p(s'|s,a)$ which specifies the probability of transitioning to state $s'$ from state $s$ and executing action $a$. The policy function $\pi_\theta(a|s)$, parameterized by $\theta$, specifies the distribution over actions $a$ given a state $s$. $\rho^\pi(s)$ denotes the stationary distribution over the state space given that policy $\pi$ is followed. We denote the total discounted reward as $r^\gamma_t = \sum^\infty_{i=t} \gamma^{i-t} r(s_i,a_i)$, where $\gamma \in [0.0, 1.0]$ is the discount factor. The value function is $V^\pi(s) = \mathbb{E}[r_1^\gamma | S_1 = s, \pi]$ and the state-action value function is $Q^\pi(s, a) = \mathbb{E}[r_1^\gamma | S_1 = s, A_1=a,\pi]$. Reinforcement learning (RL) consists of class of algorithms which can be used to find the optimal policy for MDPs~\cite{SuttonB98}. RL algorithms seek to find the policy (via $\theta$) which maximizes the average expected total discount reward.

\subsection{Policy Gradients}\vspace{-0.05in}
Policy gradient methods directly optimize the policy parameters $\theta$ and are a popular type of algorithm for continuous control. They directly optimize the expected average reward function by finding the gradient of the policy function parameters. The objective function can be written as:
\begin{equation}
\label{eq:J}
J(\pi_{\theta})  = \int_{S}\rho^{\pi}(s) \int_A \pi_\theta(a|s) Q^\pi(a,s) da ds
\end{equation}
\cite{SchulmanMLJA15} showed that the gradient can also be in the form of:
$
\nabla_\theta J = \mathbb{E}[\nabla_\theta \log \pi_\theta(a|s)A(s,a)],
$
where $A$ is the advantage function. $A$ can also be replaced by other terms such as the total return, which leads to the REINFORCE algorithm~\cite{Williams92}. In addition, based on the Policy Gradient Theorem, a widely used architecture known as the \emph{actor-critic} replaces $A$ with a critic function, which can also be learned via temporal difference and guides the training of the actor, or the policy function~\cite{SuttonMSM00}.

\subsection{Proximal Policy Optimization}\vspace{-0.05in}
Proximal Policy Optimization (PPO)~\cite{schulman2017proximal} is an online policy gradient method which minimizes a new surrogate objective function using stochastic gradient descent. Compared to traditional policy gradient algorithms where one typically performs one gradient update per data sample, PPO is able to achieve more stable results at lower sample complexity. The surrogate objective is maximized while penalizing large changes to the policy. Let $r_t(\theta)$ be the ratio of probability of the new policy and the old policy: $r_t(\theta) = \frac{\pi(a_t|s_t)}{\pi_{old}(a_t|s_t)}$, then PPO optimizes the objective:
\begin{equation}\label{eq:ppo}
L(\theta) = \hat{E}_t \Big [ min( r_t(\theta) \hat{A}_t, clip( r_t(\theta), 1-\epsilon, 1+\epsilon) \hat{A}_t ) \Big ]
\end{equation}
where $A_t$ is the estimated advantage function and $\epsilon$ is a hyperparameter (e.g. $\epsilon=0.2$). The algorithm alternates between
sampling multiple trajectories from the policy and performing several epochs of stochastic gradient descent (SGD) on the
sampled dataset to optimize this surrogate objective. Since the state value function is also simultaneously
approximated, the error for the value function approximation is also added to the surrogate
objective to form the overall objective.

Online model-free algorithms such as PPO do not make use of the experience replay buffer and it is beneficial to collect experiences in parallel to reduce the variance of the gradient updates. This can be easily done by simultaneously launching $N$ agents in parallel and collect the experiences into a minibatch and update using the average gradient. While we found that PPO fairly stable and efficient, there are certainly various other alternative algorithms that one could use instead~\cite{LillicrapHPHETS15,HorganEtAl2018,EspeholtEtAl2018}. We leave experimenting with different underlying deep RL learning algorithms to future work.

\begin{figure}[t]
\begin{center}
\includegraphics[width=1\linewidth]{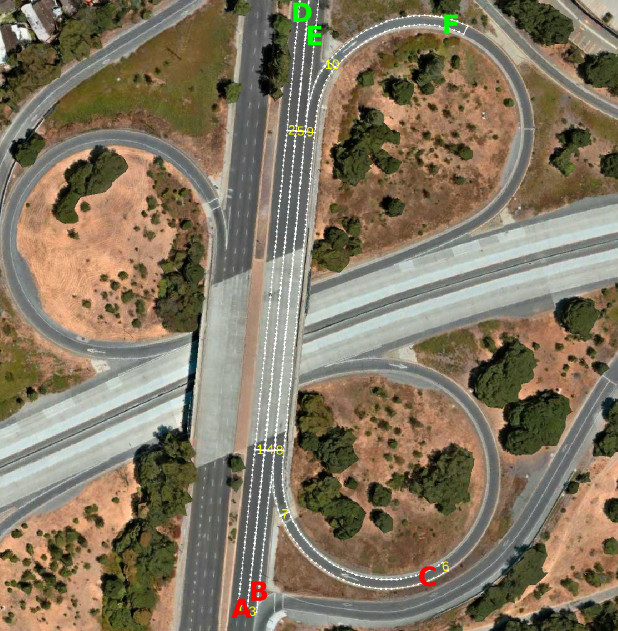}
\end{center}
\vspace{-0.1in}
\caption{\textit{Multi-agent zipper-merge simulation environment. Agents are randomly spawned at lane segments ({\color{red} A}, {\color{red} B}, or {\color{red} C}) with a goal of arriving at a randomly chosen goal location ({\color{green} D}, {\color{green} E}, or {\color{green} F}). Agents must also obey traffic laws and stay on the road}.}
\label{fig:zm_env}
\vspace{-0.1in}
\end{figure}

\begin{table*}
\small
  \begin{tabular}{ccccccccc}
    \toprule
Scale (m) & Init vel. & IDM desired vel. & \# Other agents & Spawn pr. &  $R_{success}$ & $R_{collision}$ & $R_{out-of-bounds}$ & $R_{velocity}$\\
    \midrule
	340 & [0, 5](m/s) & [10, 20] (m/s) & [0, 10] & $1\%$ &  $100.0$ & $-500.0$ & $-250.0$ & $0.1$ \\
    \midrule
  \bottomrule
\end{tabular}
\vspace{-0.1in}
\caption{ \textit{Zipper merge environment details. Bracketed values are the [min, max] bounds from which we randomly sample.}}
\label{tab:driving_env}
\end{table*}

\section{Environment}\label{sec:environments}\vspace{-0.05in}
We begin by first describing our simulation environment, as all of the subsequent discussions can be better understood in the context of the environmental details. Fig.~\ref{fig:zm_env} shows our multi-agent zipper merge RL environment, where agents start at one of (A, B, C) locations with the goal of getting to one of (D, E, F) destinations. Both starting and destination locations are chosen at random at the start of each episode. The action space of the environment consists of acceleration, steering, and turn signals.

Zipper merges, also known as \emph{double merges}, are commonly recognized as a very challenging scenario for autonomous agents~\cite{shalev2016safe}.
It is challenging as the some of the agents on the left lane intend to merge right while most of the right lane agents need to merge left. Signals and subtle cues are used to negotiate who goes first and which gap is filled. The planning also has to be done in a short amount of time and and within a short distance.

To capture the geometry of real roads, we used a satellite image as a reference and annotated polyline-based lanes to match approximately the different lanes and curvature of the road (white lines in Fig.~\ref{fig:zm_env}). Our simulation is in 2D, which is a reasonable assumption to make as we are more interested in high-level negotiations as opposed to low-level dynamic vehicle control. The discretization of our simulation is at 100 milliseconds and the frame rate is 10 Hz. The state of every agent is updated in a synchronous manner. Tab.~\ref{tab:driving_env} details the parameters of our learning environment.

\subsection{Dynamics}\vspace{-0.05in}
We use the discrete time Kinematics Bicycle model~\cite{KongPSB15} to model vehicle dynamics in our environments. The non-linear continuous time equations that describe the dynamics in an inertial frame are given by:
\vspace{-0.1in}
\begingroup
\begin{align}
\dot{x} &= v \cos (\psi + \beta), \ \ \ \  \dot{y} = v \sin (\psi + \beta)\\
\dot{\psi} &= \frac{v}{l_r} \sin (\beta), \ \ \ \ \ \ \ \ \ \ \dot{v} = a
\end{align}
\endgroup

where $x$, $y$ are the coordinate of the center-of-mass of the vehicle, $\psi$ is the inertial heading and $v$ is the velocity of the vehicle. $l_r$ and $l_f$ are the distance of the center of the mass to the front and the rear axles. $\beta$ is the angle of the current velocity relative to longitudinal axis of the vehicle. In an effort to be realistic, we bound the acceleration to be between $[-6$ to $4]$ $m/s^2$.

\subsection{Road Network}\vspace{-0.05in}
The road network of our environment consists of a 2D graph of straight, curved, and polyline lanes. Straight lanes are modeled by a left boundary line and a right boundary line. Curved lanes are modeled using Clothoids~\cite{Marzbani15} with a specified lane width. Finally, arbitrary shaped lanes are modeled by separate left and right lane boundaries. In our 2D graph, each lane is a node is this graph and connected by incoming and outgoing edges. The direction of the edge indicates the direction of travel.

\subsection{IDM Agents}\vspace{-0.05in}
As part of the environment, we have rule-based agents of differing ``intelligence". Also known as intelligent driving models (IDMs)~\cite{kesting2010enhanced}, the simplest agents perform lane-keeping starting from a specified lane using adaptive cruise control (ACC): slowing down and speeding up accordingly with respect to the vehicle in front. Building on top of these simple ACC agents, we add lane change functionality by using gap acceptance methods~\cite{hwang2005modeling}, so that merge can be done in a safer manner. We also vary the desired and starting velocities and accelerations to create a population of different IDM agents (they will drive a little differently from each other). To differentiate between the IDM agents, we will refer to the agent/vehicle that we are learning to control as the \emph{ego} vehicle.

\section{Model Architecture}\label{sec:architecture}\vspace{-0.05in}

In this section, we describe in detail our deep RL framework for self-play, which is used to learn sensible negotiation policies in the aforementioned zipper merge environment. We will explain the architecture of the policy network, the observation space and the cost function used for learning. At the core, our learning is a distributed form of the PPO policy gradient algorithm, where multiple environments are simulated in parallel to collect experiences. After first training a single deep RL agent with an environment populated with other rule-based IDM agents, we initiate self-play by replacing a portion of the agents with previously learned RL agents. As self-play training progresses, we iteratively add more RL agents with updated parameters to the training agent population.

\subsection{Observations and State Encodings}\vspace{-0.05in}
The road geometry, state of other agents, and the goal/destination of the ego vehicle are all important for decision making. The observations from our RL agent's perspective consist of a combination of two components. The first is a top down rasterization view in ego-centric frame: see examples of this rasterization in Fig. \ref{fig:network_diag} (top left). This rasterization captures the lane geometry, the route information, and the poses of all vehicles, essentially capturing the context of the scene. The rasterization is an RGB image of $128 \times 128 \times 3$. Rasterization is a simple encoding scheme that renders the scene into a sequences of image frames. The rasterized input contains all of the geometric relationships in metric space. As we will discuss later, state encoding will then be extracted by a convolutional neural network (CNN).  We use OpenGL for efficient rasterization. Besides rasterization, we also use a 76 dimensional vector which is composed of the position, velocity, acceleration, orientation, and turn signal states of the 8 closest neighboring vehicles to ego. In the real world, these inputs can be given by the output of object tracking. We also encode ego's ideal reference route and provide it as a part of the 76 dimensional vector. Temporal history of two frames (200 ms) is used to provide temporal contextual information.

\subsection{Policy Network}\vspace{-0.05in}
\begin{figure}[t]
\begin{center}
\includegraphics[width=0.99\linewidth]{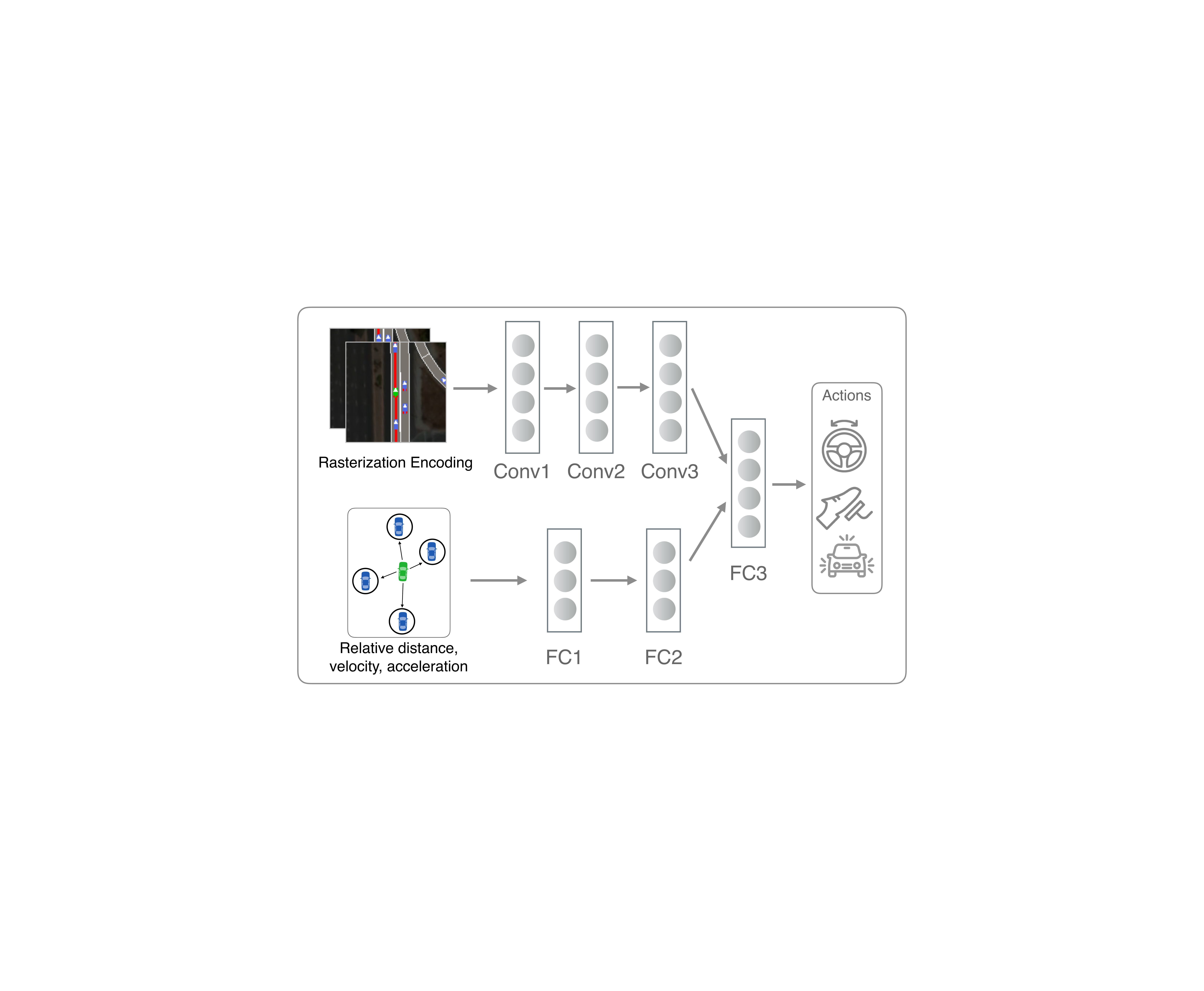}
\end{center}
\vspace{-0.2in}
\caption{\textit{Architecture diagram of the policy function. Inputs consist of two types. The first is a rasterized top-down image of the scene, with the ego vehicle in the center. The second is the low-dimensional precise measurements of relative distances, velocities and accelerations of other vehicles with respect to the ego vehicle. These inputs are passed through convolutional and fully connected layers to generate three actions: steering, acceleration and turn signals}.}
\label{fig:network_diag}
\end{figure}

The policy function (Fig.~\ref{fig:network_diag}) is represented by a deep neural network (CNN) which takes the state encoding of the environment (vehicle poses, lanes, etc.) as input and outputs steering, acceleration and turn signal commands for the ego vehicle. Our policy function is ``end-to-end" in that it bypasses traditional trajectory planning and feedback control. Our policy has two streams of inputs, which are combined via late fusion at the third fully-connected (FC3) layer. The two streams are complementary to each other as rasterization provides spatial context but it is discretized and lossy. The low dimensional relative measurements are highly precise but do not capture road information. A 3-layer convolution neural network is used to process the rasterization stream. The layers of the CNN have a receptive field size of 4 with a stride of 2 to process the input channels. After 3 convolution layers, a fully connected layer embeds the scene into a $128$ dimensional vector. In the other stream, two fully connected layers of dimension $64$ encode the $76$ dimensional input. The output of the two streams are concatenated into a single layer and an additional fully connected hidden layer is used to predict the actions. The action space is 3 dimensional: steering, acceleration, and turn signals.

\subsection{Reward Function}\vspace{-0.05in}
The reward function, or cost function, is critical for guiding the search for the desired optimal policy. We define collision as the intersection of the bounding boxes of two vehicles, with a $-500.0$ reward. A vehicle goes out-of-bounds when its centroid is more than $0.75 \times$ the lane width from the center-line of any lane. The cost incurred for out-of-bounds is $-250.0$. Successfully completing the zipper merge gains a $+100.0$ reward. There are also penalties of $-0.1$ whenever the turn signal is turned on to prevent the policy from constantly having the turn signal lights on. Vehicle velocity up to the speed of $15$ m/s is rewarded with a scaling of $0.1$. In order to reward a vehicle for remaining in the center of its lane, a lane center offset penalty of $-0.1$ is applied for every meter off the center line. Finally, we also penalize for non-smooth motion by a penalty of $-2.0$ multiplied with the change in steering angle. 

It is also critical to introduce reward shaping~\cite{ng1999policy} to make learning easier. Specifically, we linearly anneal the crash and out-of-bounds loss from $-100.0$ to their final value of $-500.0$ and $-250.0$, respectively, as training progress from $0$ to $1000$ updates. These hyperparameters are selected manually to ensure sensible performance. A better approach would be to find these values from real data, we leave this to future work.

\subsection{Training}\vspace{-0.05in}
We began by training a single RL policy using PPO with a batch size of $32$, learning rate of $0.0025$, entropy coefficient of $0.001$, and the number of environment steps between parameter updates is $1024$. Distributional learning uses $32$ separate processes to step through $32$ environments in parallel. For each episode, roughly up to $10$ agents are launched, each having their own random destination location. An episode is terminated when either one of several situations occurs: destination is reached, collision, ego going out of bounds, or after $1000$ timesteps.

\subsection{Multi-agent Self-play}\vspace{-0.05in}
\label{sec:selfplay}
\begin{figure}[h]
\begin{center}
\includegraphics[width=0.9\linewidth]{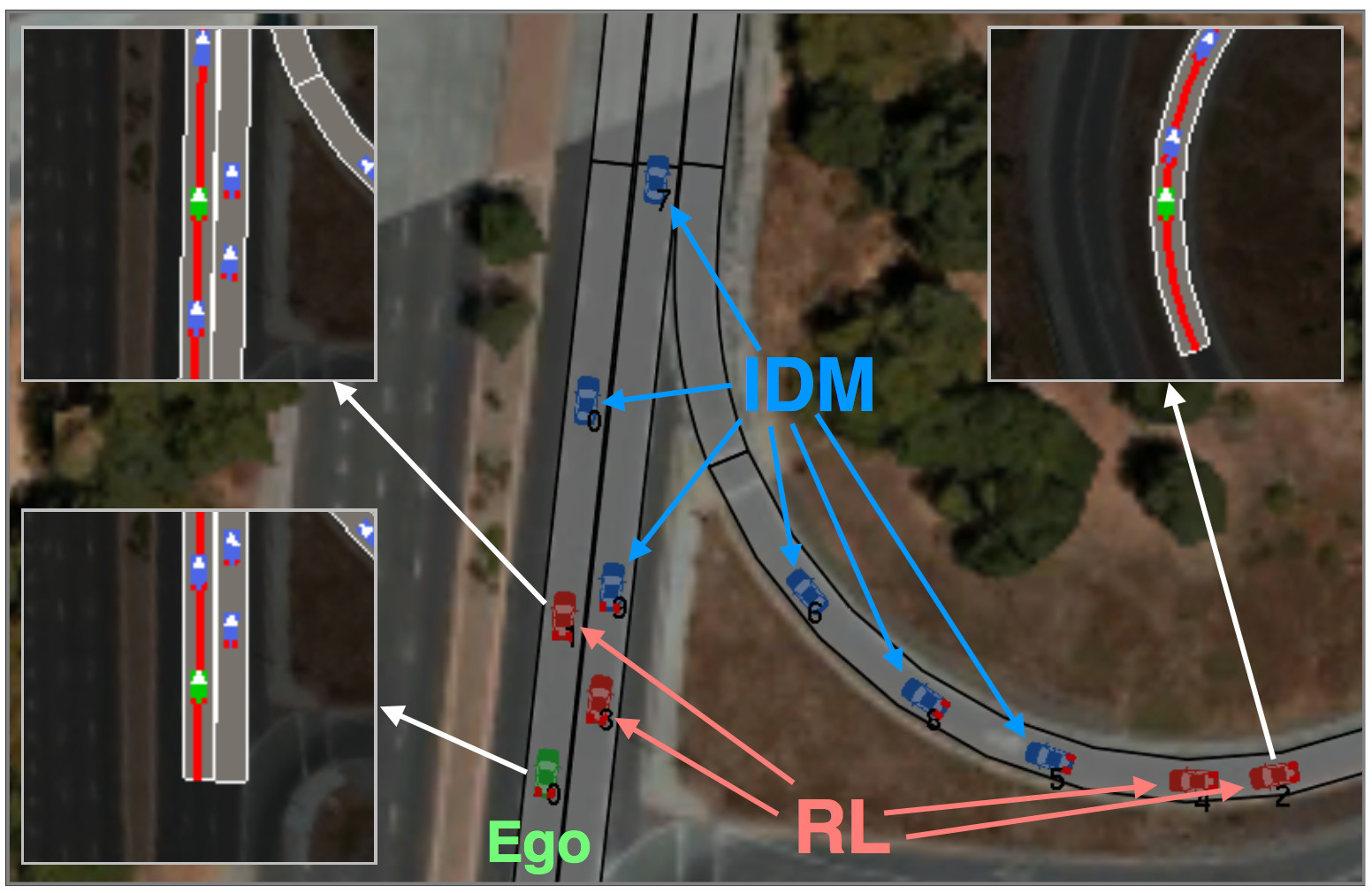}
\end{center}
\vspace{-0.2in}
\caption{\textit{Training during self-play. For ego (green), the sparring agent population consists of two types of agents, the IDM agents (blue) and other RL agents (red)}.}
\label{fig:self_play_diag}
\end{figure}

Reinforcement learned policies are well known for their ability to exploit the training environment. In the case of our multi-agent environment, it is prone to overfit to the specific behaviors of the ``sparring" IDM agents. For example, if the IDM agents have a tendency to brake suddenly and always yield to another merging vehicle, then the RL policy will learn to exploit this by being ultra aggressive. However, this could be dangerous in environments with agents that do not always yield. Additionally, it is suboptimal to train only with IDM agents as it is hard to manually tune IDM agents' parameters. For example, we noticed that IDM agents are to blame in most collisions involving IDM and RL agents.

Self-play is a very broad and general technique for increasing the complexity of the environment and learning better policies~\cite{BansalPSSM17,SilverZero18,heinrich2016deep}. At a high level, self-play iteratively adds the latest of the learned policies back to the environment, increasing the diversity and perhaps realism of the environment. For our application, starting from a simple road network and a handful of rule-based IDM agents, self-play allows us to learn complex policies capable of a diverse set of behaviors.

Fig.~\ref{fig:self_play_diag} shows a scene of self-play training with a mixed set of agents. Rule-based IDM agents are in blue while the RL agents are in red. The green ego agent is also an RL agent. The ego and RL agents share the same policy, but their input representation is normalized to be the center of the world (see the insets pointed to by the white arrows). For each episode, there is a mix of IDM and RL agents with different policy parameters.

We divide our self-play training in 3 stages. In the 1st stage, the RL policy is trained in the \emph{sole} presence of rule-based IDM agents. In stage 2, self-play is trained in the presence of 30\% IDM agents, 30\% RL agents from stage 1, and the other 40\% are controlled by the current learning policy. In stage 3, we additionally train with the agents from stage 2. See Tab.~\ref{tab:self_play_popu} for the percentage\footnote{These numbers are not exact as vehicles are spawned continuously and would-be vehicles occupying the same regions are not spawned.} of agents in different populations. Alg.~\ref{alg:self_play} further details the learning algorithm.

\begin{table}[t]
  \caption{ \textit{Training agent population distribution of agent types for different self-play training and testing stages.}}
\makebox[0.5\textwidth][c]{
    \begin{tabular}{cccccc}
\toprule
& \multicolumn{4}{c}{Population Agent Types} \\
 \cmidrule(lr){2-5}
Population & IDM   & RL  & SP1 & SP2   \\
\midrule
Popul. 1   &	100\%	& 0\%	& 0\%   & 0\%  \\
Popul. 2   &	50\%		& 50\%	& 0\%  & 0\%  \\
Popul. 3   &	30\%		& 30\%	& 40\%  & 0\%  \\
Popul. 4   &	10\%		& 20\%	& 30\%	& 40\% \\
\midrule
\bottomrule
\end{tabular}%
}
\label{tab:self_play_popu}
\vspace{-0.1in}
\end{table}%

\begin{algorithm}[htp]
\small
\DontPrintSemicolon
\SetAlgoLined
\SetKwInput{Input}{Input~}
\SetKwInput{Initialize}{Initialize~}
\Input{~Environment $\mathcal{E}$, $S$ stages, $N_{idm}$ IDM agents, $N_{rl}$ self-play agents.}
\Initialize{Randomly initialize policy $\pi$.}
\For{$stage \ s = 1 \mbox{ to } S$}{
  Choose the \% mix of agents to create population, (see Tab.~\ref{tab:self_play_popu}). \;
  Reset all envs with the new agent population. \;
  set $I$ updates for this stage. \;
  \For{$i = 0 \mbox{ to } I$}{
	Launch $K$ parallel threads to gather experiences $\lbrace s_t,a_t,r_t,s_{t+1}, \dots, s_{\tau} \rbrace_k$ from all $K$ envs $\mathcal{E}_k$. \;
	Batch $ \mathcal{B} \leftarrow \lbrace s_t,a_t,r_t,s_{t+1}, \dots, s_{\tau} \rbrace_k$ for all $K$. \;
	Update policy network parameters by training on data $\mathcal{B}$ (using Eq.~\ref{eq:ppo}). \;
	}
}
\caption{Self-play multi-agent RL training}
\label{alg:self_play}
\end{algorithm}

\begingroup
\setlength{\tabcolsep}{4pt} 
\renewcommand{\arraystretch}{1.0} 

\begin{table*}[t!]
  \caption{\textit{Quantitative performance (successful completion  and collisions) of various agents against different population of agents. Reported numbers are in success percentage (collision percentage). SP1/2 denote the self-play trained agents}.}
\makebox[0.5\textwidth][c]{
\small
    \begin{tabular}{lllllll}
\toprule
 & & \multicolumn{1}{c}{Training} & \multicolumn{4}{c}{Testing Success rate \% ( Collision rate \%)} \\
 \cmidrule(lr){3-3}  \cmidrule(lr){4-7}
Stage & \multicolumn{1}{c}{Agents} & Train w/ Population & Test w/ Popul. 1     & Popul. 2  & Popul. 3 & Popul. 4  \\
\midrule
0	& IDM   & N/A							& \sucr{62.8}{3.1}{0.8}{0.6} & -         & -           & -			 \\
1 & RL    & (Popul. 1) IDM  					& \sucr{94.8}{1.4}{4.0}{1.2}	 & \sucr{77.2}{2.7}{12.4}{2.1} & -   	  	  	  & -  		  \\
2	& SP1  	& (Popul. 3) IDM+RL+SP1  		& \sucr{96.0}{1.2}{3.6}{1.2}	 & \sucr{91.2}{1.8}{4.4}{1.3}  & \sucr{95.2}{1.4}{3.2}{1.1}  & - 		  \\
3 & SP2   	& (Popul. 4) IDM+RL+SP1+SP2    	& \sucr{93.2}{1.6}{6.0}{1.5}	 & \sucr{94.8}{1.4}{4.8}{1.4}  & \sucr{95.2}{1.4}{4.4}{1.3}  & \sucr{98.2}{0.8}{1.4}{0.7} \\
\bottomrule
\end{tabular}%
}
\vspace{-0.1in}
\label{tab:results}
\end{table*}

\endgroup

\section{Experimental Results}\label{sec:experiments}\vspace{-0.05in}
We perform experiments with the aforementioned learning environment and deep reinforcement learning using a distributed learning system which simultaneously stepped through environments in parallel to collect experiences. An NVIDIA Titan X GPU is used to accelerate the learning of the policy function. Training is performed over roughly $10M$ environment update steps, which corresponds to about $278$ hours of real time experience.

\subsection{Quantitative Results}\label{sec:exp_quant}
\begin{figure}[t]
\begin{center}
\hspace{-0.2in}
\includegraphics[width=1.05\linewidth]{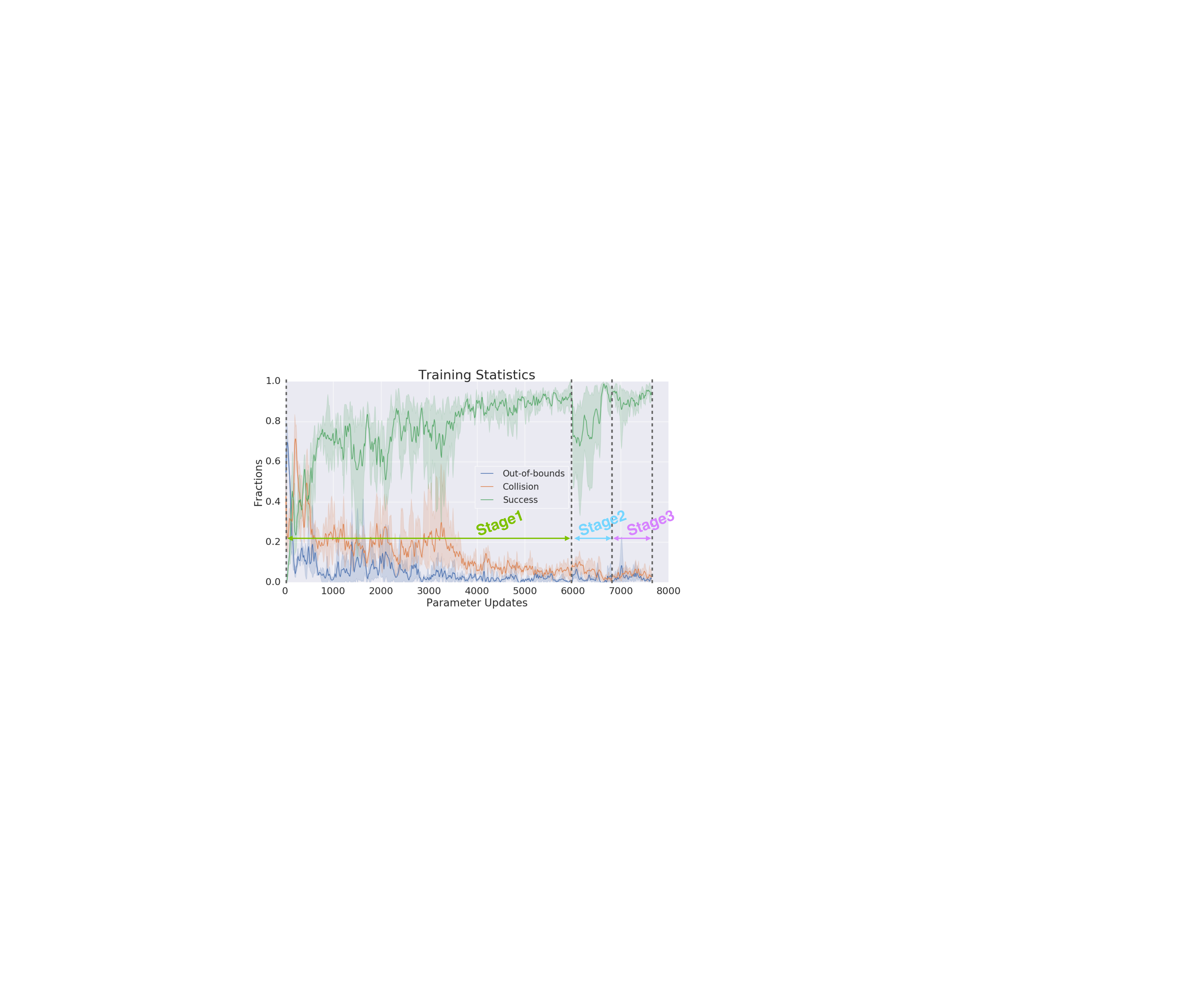}
\end{center}
\vspace{-0.25in}
\caption{\textit{Different statistics as a function of parameter updates: success rate, collision rate, and out-of-bounds rate. There are 1024 environment steps between weight updates. Arrows denote self-play stages (Sec.~\ref{sec:selfplay})}.}
\label{fig:train_stats}
\end{figure}

\vspace{-0.05in}
We plot performance measure in terms of the percentage of success\footnote{A successful merge is when an agent arrives at its desired destination location, randomly chosen at the start of the episode.}, collision rates, and out-of-bounds rates during training. Fig.~\ref{fig:train_stats} shows these statistics with respect to accumulated parameter updates across all three stages. Results are averaged over 4 random trials. Performance dips at the start of every stage as new agents are introduced. By the end of stage 3, performance peaks against a diverse set of sparring agents. 

In Tab.~\ref{tab:results}, we quantify various testing success and collision rates for policies learned at different stages. The evaluation, with standard errors, is over $250$ random trials without adding exploration noise. As we can see from Tab.~\ref{tab:results}, the success rate of the IDM agents is very poor at $63\%$. This is mainly due to the fact that they are governed by rules and gap acceptance theory~\cite{hwang2005modeling}, which is too simplistic. 
Next, in row two, we can see that RL trained agents can dramatically improves success rates ($95\%$) against the population of IDM agents. However, performance decreases to ($77\%$) when tested against a mix of IDM and RL agents, demonstrating severe overfitting.
Using three stages of self-play, we can improve the success rate up to \textbf{98\%} against a diverse population of other agents: IDMs, RL, Self-Play1, and Self-Play2 agents\footnote{See Tab.~\ref{tab:self_play_popu} for different training and testing population agent types.}.

\subsection{Qualitative Results}\label{sec:exp_quali}
\begin{figure*}[h!t]
    \begin{subfigure}[c]{0.16\textwidth}
        \includegraphics[width=\textwidth]{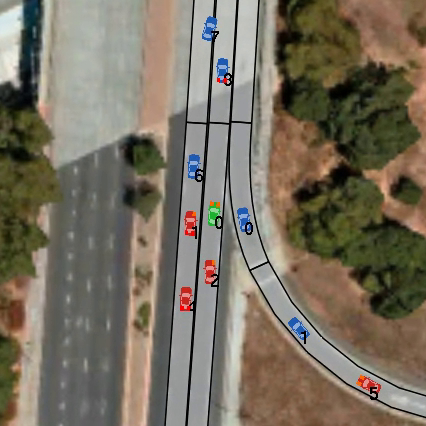}
        \caption{t=2.0 s}
    \end{subfigure}
    \begin{subfigure}[c]{0.16\textwidth}
        \includegraphics[width=\textwidth]{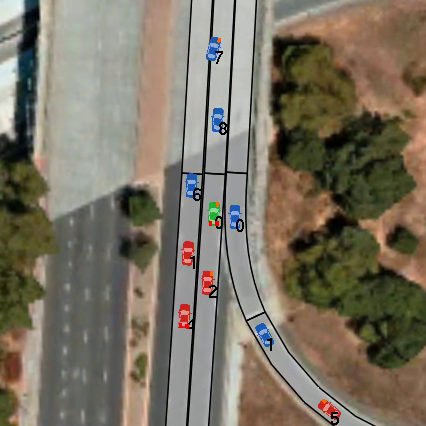}
        \caption{t=2.7 s}
    \end{subfigure}
    \begin{subfigure}[c]{0.16\textwidth}
        \includegraphics[width=\textwidth]{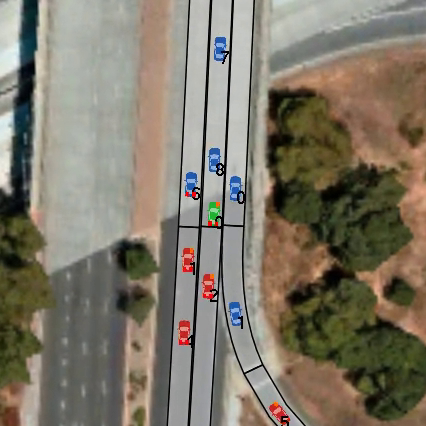}
         \caption{t=3.5 s}
    \end{subfigure}
    \begin{subfigure}[c]{0.16\textwidth}
        \includegraphics[width=\textwidth]{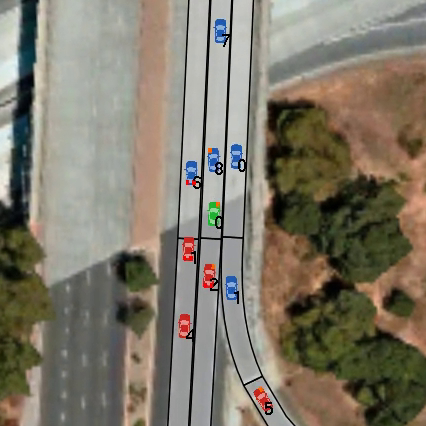}
        \caption{t=4.0 s}
    \end{subfigure}
        \begin{subfigure}[c]{0.16\textwidth}
        \includegraphics[width=\textwidth]{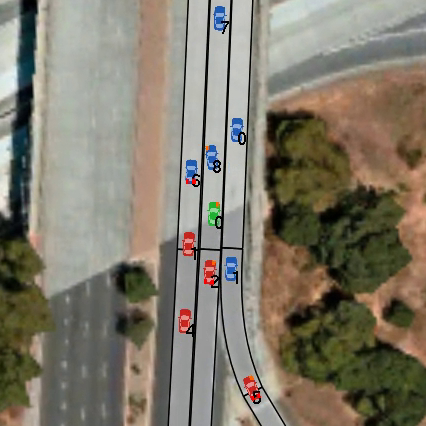}
        \caption{t=4.5 s}
    \end{subfigure}
    \begin{subfigure}[c]{0.16\textwidth}
        \includegraphics[width=\textwidth]{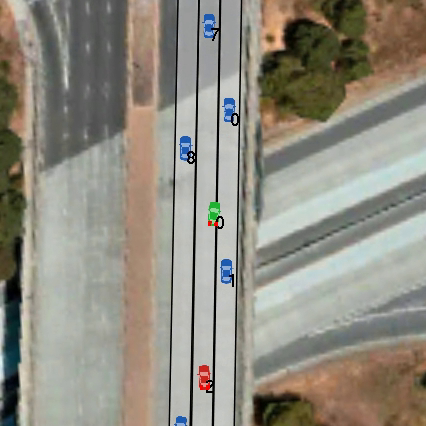}
        \caption{t=9.0 s}
    \end{subfigure}
    \vspace{-0.1in}
    \caption{ \textit{\textbf{Turn Signal + Wait and Merge}: Ego learned to use its turn signal and waited until the adjacent vehicle in the entrance lane accelerated before merging}.}
    \label{fig:behavior_tsl_merge}
\end{figure*}

\begin{figure*}[h!t]
    \begin{subfigure}[c]{0.16\textwidth}
        \includegraphics[width=\textwidth]{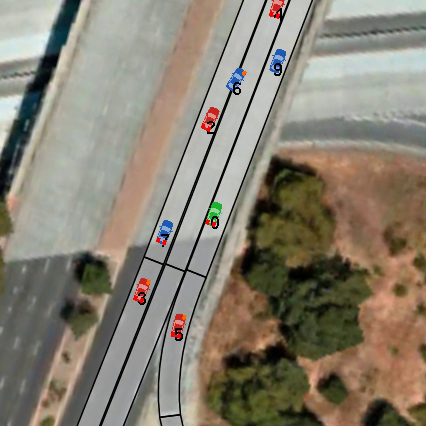}
        \caption{t=0.0 s}
    \end{subfigure}
    \begin{subfigure}[c]{0.16\textwidth}
        \includegraphics[width=\textwidth]{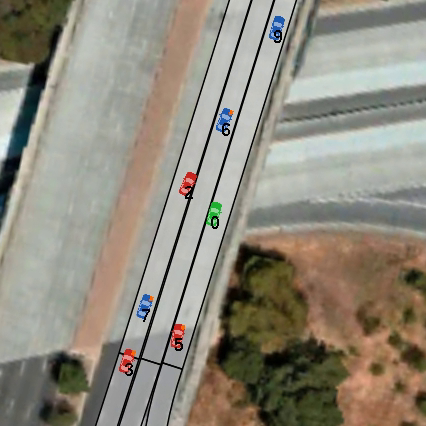}
        \caption{t=4.8 s}
    \end{subfigure}
    \begin{subfigure}[c]{0.16\textwidth}
        \includegraphics[width=\textwidth]{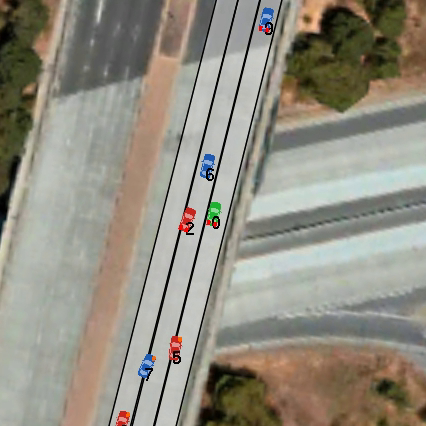}
        \caption{t=6.0 s}
    \end{subfigure}
    \begin{subfigure}[c]{0.16\textwidth}
        \includegraphics[width=\textwidth]{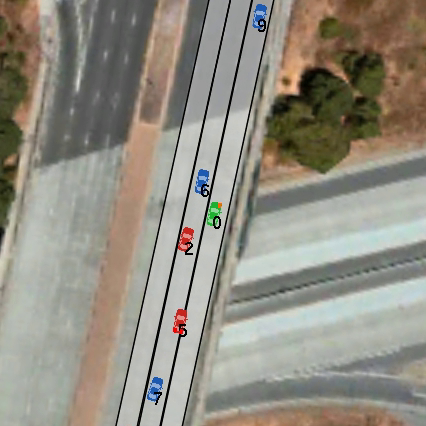}
        \caption{t=6.4 s}
    \end{subfigure}
        \begin{subfigure}[c]{0.16\textwidth}
        \includegraphics[width=\textwidth]{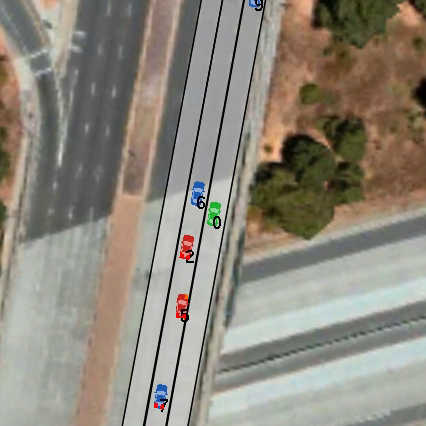}
        \caption{t=6.8 s}
    \end{subfigure}
    \begin{subfigure}[c]{0.16\textwidth}
        \includegraphics[width=\textwidth]{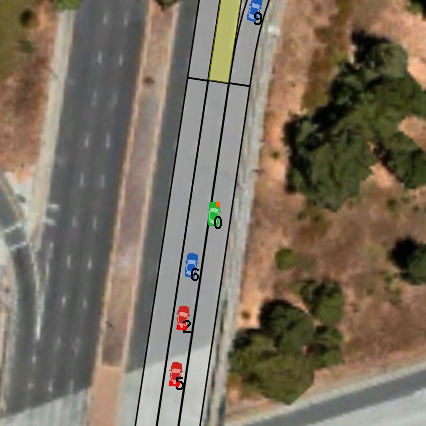}
        \caption{t=7.2 s}
    \end{subfigure}
    \vspace{-0.1in}
    \caption{\textit{ \textbf{Overtaking and finding a gap}: Ego cannot safely merge due to another vehicle merging to the right from the left-most lane at the same time. Instead, it accelerates to overtake and then merges}.}
    \label{fig:behavior_find_gap}
\end{figure*}

\begin{figure*}[h!t]
    \begin{subfigure}[c]{0.16\textwidth}
        \includegraphics[width=\textwidth]{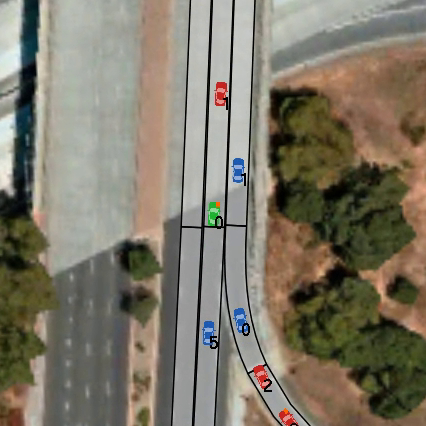}
        \caption{t=0.0 s}
    \end{subfigure}
    \begin{subfigure}[c]{0.16\textwidth}
        \includegraphics[width=\textwidth]{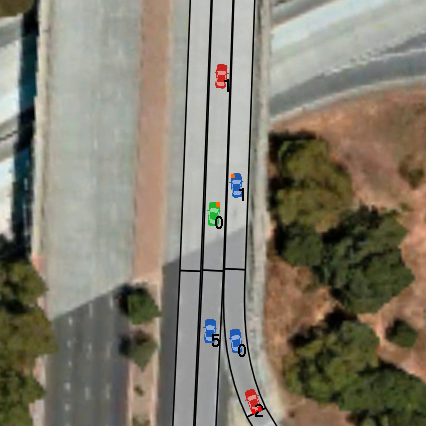}
        \caption{t=4.8 s}
    \end{subfigure}
    \begin{subfigure}[c]{0.16\textwidth}
        \includegraphics[width=\textwidth]{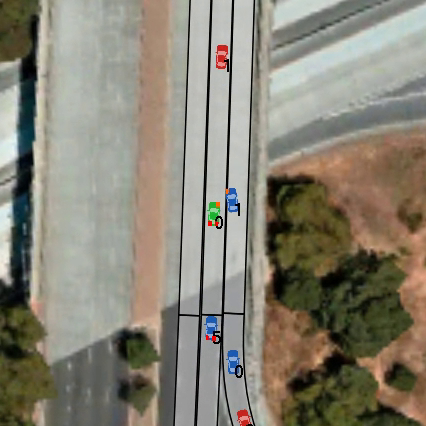}
        \caption{t=6.0 s}
    \end{subfigure}
    \begin{subfigure}[c]{0.16\textwidth}
        \includegraphics[width=\textwidth]{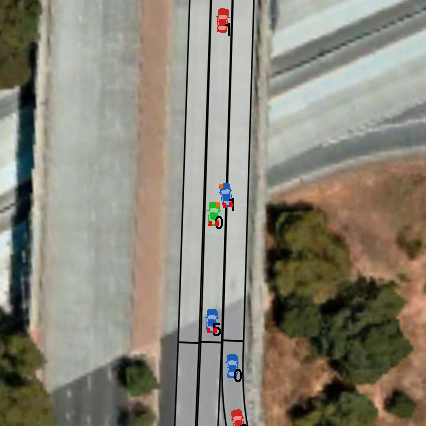}
        \caption{t=6.4 s}
    \end{subfigure}
        \begin{subfigure}[c]{0.16\textwidth}
        \includegraphics[width=\textwidth]{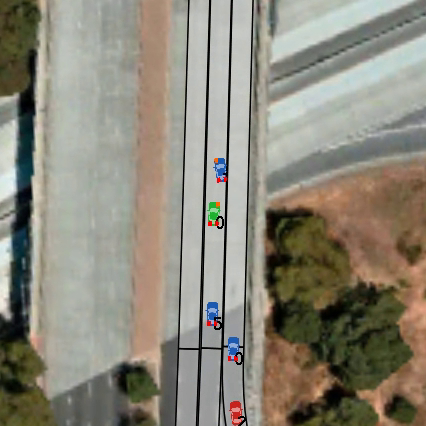}
        \caption{t=6.8 s}
    \end{subfigure}
    \begin{subfigure}[c]{0.16\textwidth}
        \includegraphics[width=\textwidth]{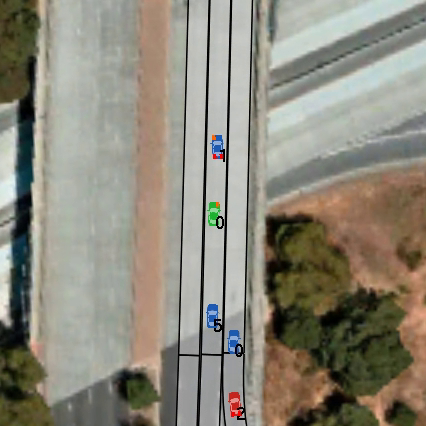}
        \caption{t=7.2 s}
    \end{subfigure}
    \vspace{-0.1in}
    \caption{\textit{\textbf{Yielding}: Ego braked suddenly to yield due to the extremely aggressive behavior of the blue IDM agent}.}
    \label{fig:behavior_yielding}
\end{figure*}

\begin{figure*}[h!t]
    \begin{subfigure}[c]{0.16\textwidth}
        \includegraphics[width=\textwidth]{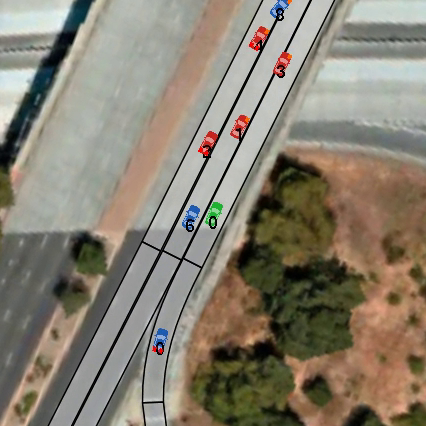}
        \caption{t=0.0 s}
    \end{subfigure}
    \begin{subfigure}[c]{0.16\textwidth}
        \includegraphics[width=\textwidth]{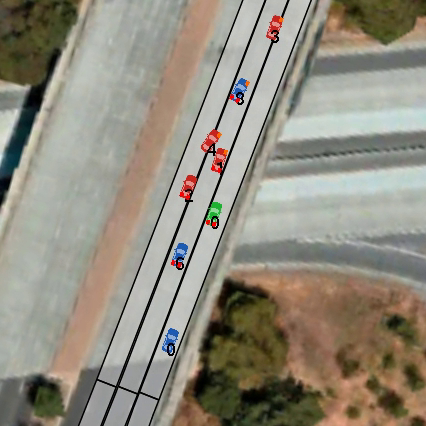}
        \caption{t=4.8 s}
    \end{subfigure}
    \begin{subfigure}[c]{0.16\textwidth}
        \includegraphics[width=\textwidth]{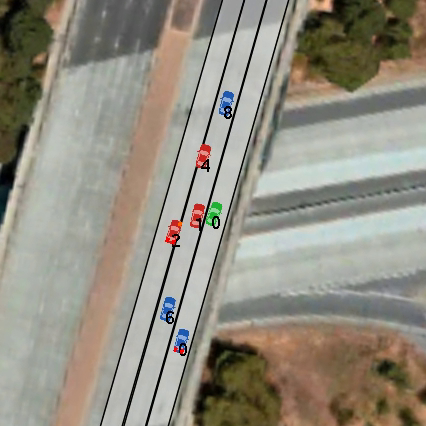}
        \caption{t=6.0 s}
    \end{subfigure}
    \begin{subfigure}[c]{0.16\textwidth}
        \includegraphics[width=\textwidth]{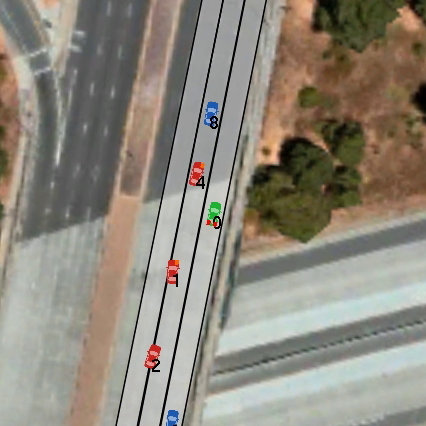}
        \caption{t=6.4 s}
    \end{subfigure}
        \begin{subfigure}[c]{0.16\textwidth}
        \includegraphics[width=\textwidth]{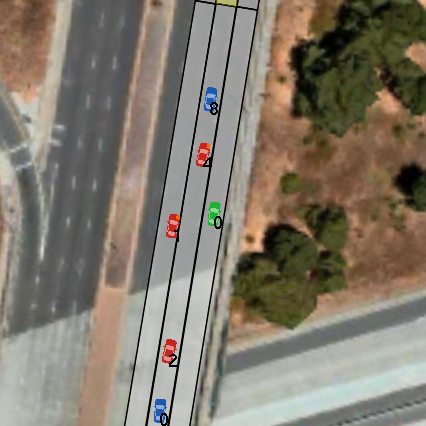}
        \caption{t=6.8 s}
    \end{subfigure}
    \begin{subfigure}[c]{0.16\textwidth}
        \includegraphics[width=\textwidth]{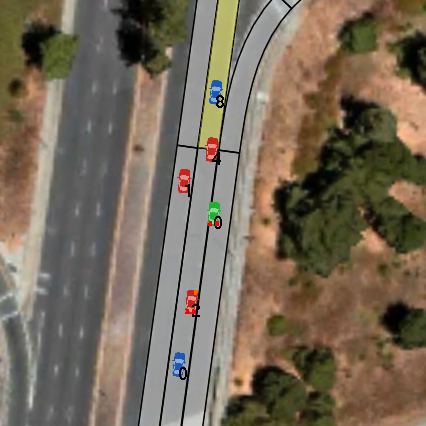}
        \caption{t=7.2 s}
    \end{subfigure}
    \vspace{-0.1in}
    \caption{ \textit{\textbf{Defensive}: Ego cautiously decides to sway back to the right portion of the lane at around the 6.0 secs mark}.}
    \label{fig:behavior_defensive}
\end{figure*}

\vspace{-0.05in}
Our policies are able to learn a variety of interesting driving behaviors with self-play. These behaviors include overtaking to merge, emergency braking, using turning signals, and defensive yielding. They showcase various multi-agent interactions and diverse behaviors exhibited by the agents during merges. The resulting policies also learned to be flexible within the confines of the road, deviating from the center of the lane when necessary. Figs.~\ref{fig:behavior_tsl_merge},~\ref{fig:behavior_find_gap},~\ref{fig:behavior_yielding},~\ref{fig:behavior_defensive} show temporal
snapshots of testing episodes. The ego agent is in green, RL and self-play agents are red, and the rule-based IDM agents are blue.

\begin{figure}[bht]
    \begin{subfigure}[c]{0.32\textwidth}
        \includegraphics[width=\textwidth]{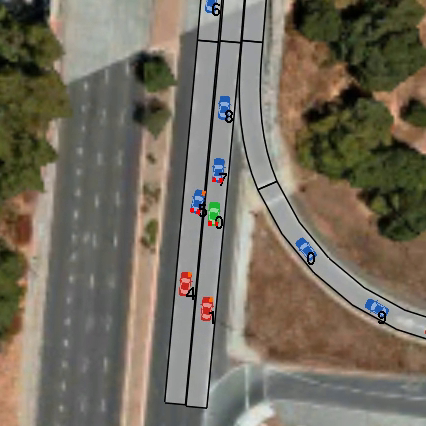}
        \caption{t=0.0s}
    \end{subfigure}
    \begin{subfigure}[c]{0.32\textwidth}
        \includegraphics[width=\textwidth]{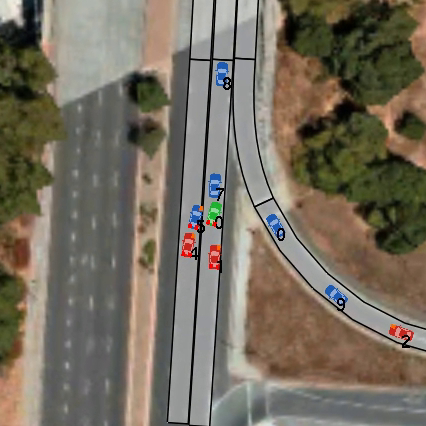}
        \caption{t=4.8s}
    \end{subfigure}
    \begin{subfigure}[c]{0.32\textwidth}
        \includegraphics[width=\textwidth]{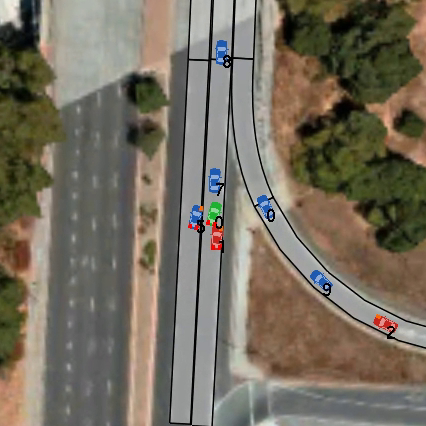}
        \caption{t=6.0s}
    \end{subfigure}
    \vspace{-0.1in}
    \caption{\textit{A failure case (rear-end). Ego (green) successfully brakes, however the vehicle behind it also brakes but not quickly enough, resulting in a rear ended collision}.}
    \label{fig:fail_crash}
\end{figure}

\begin{figure}[bht]
    \begin{subfigure}[c]{0.32\textwidth}
        \includegraphics[width=\textwidth]{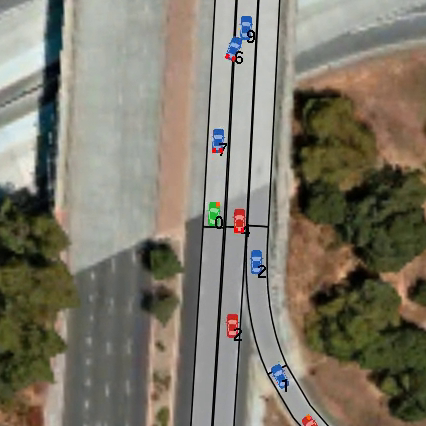}
        \caption{t=0.0 s}
    \end{subfigure}
    \begin{subfigure}[c]{0.32\textwidth}
        \includegraphics[width=\textwidth]{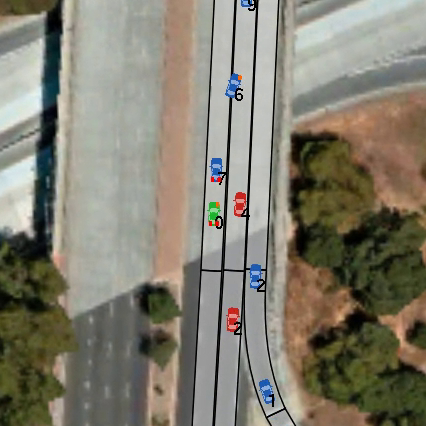}
        \caption{t=1.0 s}
    \end{subfigure}
    \begin{subfigure}[c]{0.32\textwidth}
        \includegraphics[width=\textwidth]{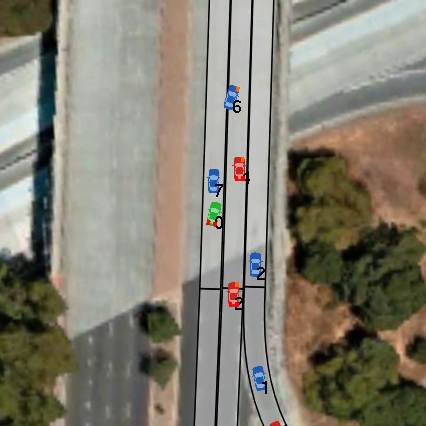}
        \caption{t=3.4 s}
    \end{subfigure}
    \vspace{-0.1in}
    \caption{ \textit{Emergency stop example. Ego vehicle (green) brakes hard due to the sudden deceleration of the vehicle in front. Instead of only braking, the policy learned to turn its wheels to the side as well, reducing chances of collisions. Note that this is done without violating the right lane}.}
    \label{fig:fail_e_stop}
\end{figure}

\subsection{Emergency and Failure Cases}\vspace{-0.05in}
Our algorithm is not perfect, whether it is due to unobservable intentions of other agents, information loss due to temporal discretization, or an inadequately scaled reward function, we still observe collisions in simulation. We analyzed some of these failure cases here. The most common cause of collisions is due to a sudden stop of the vehicle in front of ego. While the RL policy will try to brake, it is sometimes inadequate and results in a collision. In Fig.~\ref{fig:fail_crash}, while ego stops in time, unfortunately, the vehicle behind ego was unable to stop in time and caused a collision. 

Another interesting example is an emergency braking situation shown in Fig.~\ref{fig:fail_e_stop}. Here, the IDM vehicle in front of ego abruptly stops. As ego decelerates, it also steers towards the right side\footnote{We hypothesize and speculate that it was useful to increase the distance traveled the wheel until the front made contact with the vehicle in the front. Interestingly, it does not go across the right ego lane boundary.}. 
This is interesting as it loosely correlates with what is observed in some human behaviors in similar emergency breaking situations.

\section{Discussions}\label{sec:conclusions}\vspace{-0.05in}
In this paper we proposed an algorithm towards tackling the challenging problem of multi-agent robust decision making in a simulated traffic merge environment. Starting with a geometrically realistic environment but populated with only simple rule-based agents, we iteratively increased the diversity of the agent population by using self-play to add more and more capable RL-based agents. We empirically showed that three stages of self-play can dramatically increase the success rate while also keeping failure rates low. In addition, qualitatively, the learned policies exhibit some interesting and human-like behaviors. For future work, it is critical to drive the collision rates to zero. Reward shaping and modeling the future uncertainties are also two possible avenues for the future.

\vspace{0.15in}
\noindent
\textbf{Acknowledgements}
\small
\ \ We thank Barry Theobald, Hanlin Goh, Ruslan Salakhutdinov, Jian Zhang, Nitish Srivastava, Alex Druinsky, and the anonymous reviewers for making this a better manuscript.
{
\footnotesize
\bibliographystyle{ieee}
\bibliography{spsd}
}
\end{document}